\documentclass{article}

\usepackage[final,nonatbib]{neurips_2024}

\usepackage[utf8]{inputenc} 
\usepackage[T1]{fontenc}    
\usepackage{hyperref}       
\usepackage{url}            
\usepackage{booktabs}       
\usepackage{amsfonts}       
\usepackage{nicefrac}       
\usepackage{microtype}      
\usepackage{xcolor}         
\usepackage[sorting=none]{biblatex}
\usepackage{graphicx}
\usepackage{amsmath}
\usepackage{amssymb}

\usepackage{subcaption}
\usepackage{caption}
\newcommand{\mb}{\mathbf}

\title{CUAL: Continual Uncertainty-aware Active Learner}
\vspace{-4mm}

\vspace{-7mm}
\author{Amanda Rios\\
Intel\\
{\tt\small amanda.rios@intel.com}
\And
Ibrahima Ndiour\\
Intel\\
{\tt\small ibrahima.j.ndiour@intel.com}
\And
Parual Datta\\
Intel\\
{\tt\small parual.datta@intel.com}
\And
Jerry Sydir\\
Intel\\
{\tt\small jerry.sydir@intel.com}
\And
Omesh Tickoo\\
Intel\\
{\tt\small omesh.tickoo@intel.com}
\And
Nilesh Ahuja\\
Intel\\
{\tt\small nilesh.ahuja@intel.com}
}

\begin{document}

\maketitle

\vspace{-6mm}
\begin{abstract}
  \vspace{-2mm}
   
  
  AI deployed in many real-world use cases
  should be capable of adapting to novelties encountered after deployment.
  Here, we consider a challenging, under-explored and realistic continual adaptation problem: a deployed AI agent is continuously provided with unlabeled data that may contain not only unseen samples of known classes but also samples from novel (unknown) classes. 
  In such a challenging setting, it has only a tiny labeling budget to query the most informative samples to help it continuously learn. We present a comprehensive solution to this complex problem with our model "CUAL" (Continual Uncertainty-aware Active Learner). CUAL leverages an uncertainty estimation algorithm to prioritize active labeling of ambiguous (uncertain) predicted novel class samples while also simultaneously pseudo-labeling the most certain predictions of each class. Evaluations across multiple datasets, ablations, settings and backbones (e.g. ViT foundation model) demonstrate our method's effectiveness. We will release our code upon acceptance.
\end{abstract}

\vspace{-4mm}
\section{Introduction and Related Work}
\vspace{-3mm}
Real-world artificial intelligence (AI) systems often contend with evolving data distributions, due to changes in operating conditions and the introduction of new classes after deployment. To enforce a robust response, AI systems should ideally be able to continuously learn and adapt to detected novelties with minimal computing and labeling costs. 
Nonetheless, most current adaptive continual learning (CL) solutions \cite{rios2020lifelong,parisi2019continual,buzzega2021rethinking} still necessitate fully labeled data, a requirement that overlooks the feasibility and high cost of data labeling in practical AI applications. Unsupervised and semi-supervised CL solutions are scarce \cite{ordisco,Boschini2022,bagus2022supervised} and furthermore, operate with a highly simplified oracle assumption: that in unlabeled post-deployment data, old (known) classes cannot appear in conjunction with newly introduced (novel) classes. Yet, in many realistic use cases this assumption will not hold, i.e. new and old classes may co-occur. One reason current models continue to adopt this oversimplification is to limit error propagation: when unseen novel-class samples co-occur with old-classes, misclassifications can occur not only among new classes but also between novel and old classes. In this scenario, the CL model may also face prediction overconfidence \cite{ren2021survey}, equivocally mapping novel-class samples to old-class labels. The question then becomes how to design a robust semi-supervised CL model that can properly function in a non-overly constrained setting? In what is still a parallel research area, cost-effective active labeling (AL) \cite{yoo2019learning,sener2018active,gal2017deep,nguyen2022measure} has been applied extensively on closed-world learning to minimize labeling costs to a human-in-the-loop. 
However, only a few works have incorporated Active Labeling concepts within continual learning \cite{ayub2022few,vu2023active,nie2023online} as a strategy to minimize labeling costs in challenging scenarios where data distributions are evolving. 
\textbf{Our contribution is as follows}: We eliminate the commonly used oracle assumption employed by most semi-supervised CL techniques. This creates a more realistic continual learning setting where old and new classes can co-occur, while maintaining a minimal labeling budget. 
In this case, AL querying must be judiciously designed not only to prioritize novel-class sampling but also to select the most informative (ambiguous) samples among them \cite{sharma2013most}. We introduce our model, CUAL (Continual Uncertainty-aware Active Learner), which learns to classify new classes as they are introduced while efficiently utilizing the limited labeling budget (e.g., 2.5\%, an application-defined hyperparameter) by selecting only the most ambiguous samples among novel class predictions. CUAL uses its uncertainty scoring function to guide active querying. 
To our knowledge, ambiguity-based querying has not been previously covered in continual AL. 
Furthermore, unlike most existing continual AL methods that learn only from active queries \cite{ayub2022few,vu2023active,nie2023online}, CUAL leverages its uncertainty formulation to include pseudo-labeling in its AL inner loop, significantly reducing labeling requirements.
\vspace{-4mm}
\section{Our Method}
\vspace{-3mm}
\textbf{2.1. Problem Setting:} 
Consider an agent $A(x,t)$ that is expected to learn to classify from a sequence of continual tasks. At each continual task $t$, the model is presented with an initially unlabeled set of samples $U(t)$ which consists of a mixture of unseen samples of old classes $U_{old}(t)$ and unseen samples of new (novel) classes $U_{new}(t)$: 
\vspace{-3mm}
\begin{gather}
U(t) = U_{old}(t) \cup U_{new}(t), \text{where } 
U_{old}(t)  = \{x | x \thicksim \bigcup_{k=1}^{t-1} D_k\}, U_{new}(t) = \{x | x \thicksim D_t\},
\end{gather}
\vspace{-5mm}

Here $D_t$ comprises samples from the set of new classes $C^t_{new}$ introduced at task $t$, while $\bigcup_{k=1}^{t-1} D_k$ are samples belonging to all the old classes $C_{old}^{t}$ that have been learned up to and including task $t-1$. Samples in $U_{old}(t)$ are ``unseen'', meaning they were never used, neither in the initial training nor during prior tasks' learning. Note that addressing data drifts in $U_{old}$ is beyond the scope of this work and will be addressed in future research.


\textbf{2.2. Our solution:} 
We introduce a `short-term' active learner $A_s(x,i;t)$ whose goal is to acquire a sufficient number of samples from $U_{new}(t)$ via a combination of uncertainty-guided active-labeling and pseudo-labeling ($i$ being the index for inner-loop AL iterations). These will be used to update the parameters of $A(x,t)$ via continuous learning procedures such as experience replay.
Both modules, $A$ and $A_s$, contain trainable classification heads, $A^{cl}$ and $A_s^{cl}$ respectively, that operate on features extracted from a frozen large/foundation vision model pre-trained on a separate dataset with no class overlap with the test dataset (implementation details in Appendix 4.2.2). While $A^{cl}$ predicts labels for the main continual classification task, $A_s^{cl}$ will perform pseudo-labeling by predicting among only new-class labels and will use an uncertainty metric \ref{eq:uncertainty-uncanny} to guide AL and confident pseudo-labeling, as detailed in sec. 2.2.2. 
\textbf{2.2.1. Building block of CUAL's uncertainty formulation:}
CUAL's uncertainty estimation \ref{eq:uncertainty-uncanny} uses the \textit{feature reconstruction error} (FRE) metric introduced in \cite{ndiour2020probabilistic}. FRE has been shown to effectively estimate novelty in the closed-world and continuous setting \cite{rios2022incdfm}. 
In multi-class settings, FRE learns a dimensionality-reducing PCA (principal component analysis) transform $\mathcal{T}_m$ and its inverse $\mathcal{T}_m^{\dagger}$ for each class $m$. 
For scoring, a test-feature $u=g(x)$ is first transformed into a lower-dimensional subspace by applying $\mathcal{T}_m$ and then re-projected back into the original higher dimensional space 
via $\mathcal{T}_m^{\dagger}$. FRE for the $m^{\text{th}}$ class is calculated as the $\ell_2$ norm of the difference between the original and reconstructed vectors, defined in equation \ref{eq:FRE} in appendix 4.1.3. Intuitively, if a sample does not belong to the same distribution as that $m^{\text{th}}$ class, it should result in a large $FRE_m$ score. 


\textbf{2.2.2. CUAL Step by Step:}
At the outset (task $t=0$), we assume that $A(x,t=0)$ has been previously trained to classify among a fixed set of pre-deployment classes $C_{new}^0$. Simultaneously, a set of FRE transforms, $\mathcal{T}_m$ have been learnt $ \forall m \in C_{new}^0$. 
For each subsequent task $t>0$, CUAL invokes its active learning inner-loop, indexed by $i$.
At the first iteration $i=0$ of a given task $t$, a first cycle of AL is performed by querying samples with the highest uncertainty scores
defined as $S^0(u) \triangleq \min_{j\in C_{old}^t} FRE_j^0(u)$ as these are most likely to be novel (see Appendix 4.1.2 for details).  
At this point, novel classes are identified (denoted by $|C_{new}^t|$ in section 2.1, assuming $|C_{new}^t|> 0$) and the actively labeled samples are used to: (1) Initialize and train $A_s^{cl}(x,i=0,t)$ to learn an imperfect initial mapping to the $|C_{new}^t|$ novel classes ($A^{cl}$ contains output nodes only w.r.t novel classes); (2) compute rough estimates of per-novel-class PCA transforms $\{\mathcal{T}^{t,0}_m\}, m\in C_{new}^t$. Note that it's possible that not all true novel classes are found in this initial iteration and may be found in subsequent ones. For subsequent iterations $i>0$, given an unlabeled sample $x \in U(t)$, $A_s^{cl}(x,i,t)$ predicts a pseudo-label $m, m \in C_{new}^t$ which then routes the selection of the corresponding PCA transform $\mathcal{T}^{t,i-1}_m$ resulting in that iteration's uncertainty score $S^i(u)$ \ref{eq:uncertainty-uncanny}:
\vspace{-2mm}
\begin{equation}
\label{eq:uncertainty-uncanny}
    S^i(u) = \min_{j\in C_{old}^t}\frac{FRE_j^0(u)}{FRE^{i-1}_m(u)}; i>0, m=A_s^{cl}(u,i-1,t) \in C_t^{new}\\
\end{equation}
$S^i(u)$ can be used to robustly categorize samples in $U(t)$ as: \textbf{(1) Novel with high-confidence:} These are samples with the highest score values, which will occur for a high numerator relative to the denominator. A high value of numerator implies large distance from previously seen classes $C_{old}^t$, while a low value of the denominator implies low distance from novel class $m$. Such a sample likely belongs to $U_{new}(t)$ and is a strong candidate to be pseudo-labeled.
From these, we select the topmost most confident $\alpha$ percent of samples to pseudo-label \textbf{(2) Old-class with high-confidence:} lowest score values corresponding to low numerator (low distance w.r.t $C_{old}^{t-1}$) and high denominator value (high-distance from the predicted novel class $m$). Such a sample likely belongs to $U_{old}(t)$, i.e. to an old class that has already been learned; \textbf{(3) Ambiguous:} Samples for which the score is neither definitively high nor definitively low. These could be old-class samples having relatively high scores, or new-class samples having relatively low scores. Owing to this ambiguity, a clear determination cannot be made. Hence, these samples are excellent candidates for active labeling (see Appendix 4.1.2 for details).
At each inner-loop iteration, accumulated active and pseudo-labeled samples are used to re-update $A_s^{cl}(x,i,t)$ parameters. 
Finally, at the end of the inner-loop, the full set of actively labeled and pseudo-labeled samples is used to update the continual agent's classification head $A^{cl}(x,t)$ via Experience Replay \cite{rolnick2019experience}, a common CL technique. The accumulated samples are also used to compute final estimates of the PCA transforms $\{\mathcal{T}^{t,i=I}_m\}$ for $m \in C_t^{new}$. Methodology details, including those for experience replay, stopping criteria and ambiguity formulation, can be found in the appendix sections.




\textbf{CUAL Key Takeaways:} 
(1) \textbf{Ambiguity}: In continual AL, the informativeness of a sample does not necessarily equate to its distance from old-classes or to its confusion among other new classes. Rather, CUAL is, to the best of our knowledge, the first continual active learning method to characterize samples as informative if they can be classified (pseudo-labeled) as new but with \textit{ambiguous} certainty; (2) \textbf{Pseudo-labeling}: A design choice in current closed-world and also continual AL methods \cite{vu2023active,ren2021survey} is to train only with the selected labeled samples for purpose of data efficiency. However, this can come at a high cost to performance when unlabeled samples are the vast majority (tiny AL budget). This effect is amplified in continual settings as other pressures inherent to CL, e.g. catastrophic forgetting, come into the mix. There are many techniques in semi-supervised learning to leverage information of unlabeled samples. CUAL uses pseudo-labeling, a generic semi-supervised technique, to help learn within the inner loop. Yet, if wrong pseudo-labels are used to train, it can lead to severe degradation of performance. This is why CUAL's uncertainty formulation is key to ensuring that only the most confident pseudo-labels are used to enhance AL and not undermine it, as shown in Fig 2. Note that other semi-supervised techniques, such as contrastive or prediction-consistency, etc, are orthogonal to CUAL and can be integrated in future research.

\vspace{-4mm}
\section{Experiments}
\vspace{-4mm}
\textbf{3.1. Setup:} Implementations for CUAL and baselines: All use a large/foundation frozen feature extractor, e.g. ResNet50 \cite{he2016deep} pre-trained on ImageNet1K via SwAV \cite{caron2020unsupervised} or ViTs16 \cite{alexey2020image} pre-trained on Imagenet1K via DINO \cite{caron2021emerging}. $A_s^{cl}$ (the pseudo-labeling head) is a fully connected layer and $A^{cl}$'s is a perceptron of size 4096 (the latter $A_{cl}$ is same for all baselines). 
We test on 4 diverse datasets: Imagenet21K-OOD (Im21K-OOD) \cite{ridnik2021imagenet21k}, Places365-OOD (Places) \cite{zhou2017places}, Eurosat \cite{helber2019eurosat}, iNaturalist-Plants-20 (Plants) \cite{vanhorn2018inaturalist} and Cifar100-superclasses \cite{cifar100}. All of the aforementioned datasets were constructed to have class orthogonality (be out-of-distribution) with respect to Imagenet1K with the exception of Cifar100. Our replay buffer stores a fixed amount of pre-logit deep-embeddings and labels/pseudo-labels. We set the maximum coreset size to 5000 for Im21K-OOD, Places and 2500 for Cifar100, Eurosat. While Cifar100 is not orthogonal to Imagenet1K, we include it for direct comparison with our main benchmarking approach, incDFM, which uses Cifar100 Superclasses as its hardest experimental dataset. Details on datasets are in the supplementary. At each incoming unlabeled pool, we fix a mixing ratio of 2:1 of old to new classes per task, with old classes drawn from a holdout set (0.35\% of each dataset). We set pseudo-labeling selection to $\alpha=20\%$ of samples predicted as novel (appendix 4.1.1). For experiments not purposely varying the tiny supervision budget, we fix a labeling budget of 1.25\% for Im21K-OOD, Places, Plants and 0.5\% for Eurosat, Cifar100 as guided by Fig 1 \textit{right} which varies the AL budget from 0.5\% to 5\%. We compare CUAL to: (1) CCIC \cite{Boschini2022} a semi-supervised CL model that adapts MixMatch \cite{berthelot2019mixmatch} to the CL setting; (2) Experience-Replay "ER" \cite{rolnick2019experience, buzzega2021rethinking}. ER is adapted to learn with active labeling via Entropy or Margin similarly to other continual AL works \cite{vu2023active}; (3) PseudoER \cite{pseudoDER}, same as ER, but uses the AL heuristic (Entropy or margin) to iteratively pseudo-label the most confident samples during the inner loop. Other baselines are constructed (Fig 2 right table) from removing elements of CUAL such as the iterativeness (i.e. doing AL/Pseudo-labeling in one shot during ER/PseudoER), etc. 
More details in appendix sections. 

\textbf{3.2. Results:} Fig. \ref{fig:plots} (left) shows the cumulative accuracy of the continual-learning classifier module $A^{cl}$ over tasks, i.e. the average over all classes learned up to and including the task $t$. The ``Oracle'' method constitutes an upper-bound. It has perfect knowledge of old and new class labels (100\% supervision) and is trained using the same architecture and experience replay hyper-parameters. The baseline ER only uses actively labeled data (chosen to have ambiguous classification entropy values akin to CUAL, for fair comparison). While this is superior to labeling randomly (see Fig 2 left table), ER performance is still capped by the tiny labeling budget available. On the other hand, CUAL as well as semi-supervised baselines PseudoER and CCIC all employ some technique (refer to section 3.1) to learn from unlabeled samples beyond those which are actively labeled. Nevertheless, CUAL outperforms all baselines by a large margin, even with stringent labeling budgets of $1.25-2.5\%$. Also, \ref{fig:plots} (right) analyzes the effect of varying the tiny active labeling budget. We show CUAL over-performs the other methods over a large interval of supervision budgets (all tested).
Note that CCIC (blue line or blue star symbol) drastically under-performs all other approaches even with a large supervision budget of 10\%. As most semi-supervised CL methods, it originally assumed old and new classes would not co-occur and cannot properly quantify uncertainty when this assumption is lifted. Similar trends are observed in Table \ref{fig:tables} (left), which lists the continual classification performance of CUAL and baselines averaged over all tasks for all 4 datasets. We showcase results for two architecture variations: Resnet50 trained via SWAV (R50) and ViTs16 trained via DINO (ViT). 
Equivalent results for Cifar100 and other class increment variations in Appendix. 

\vspace{-3mm}
\begin{figure}[h]
\centering
\includegraphics[width=14cm]{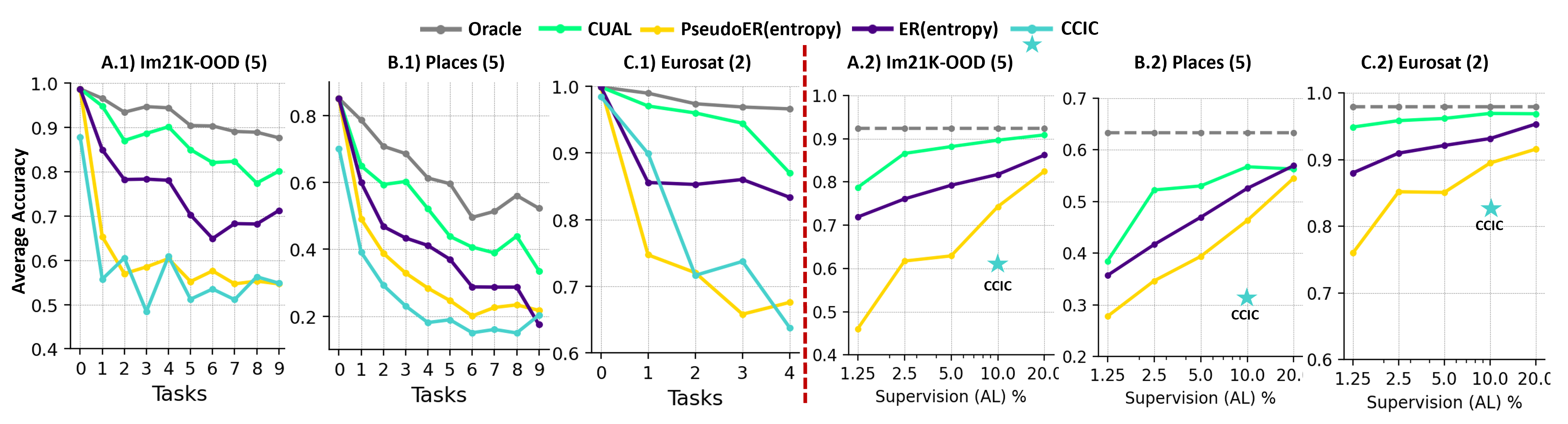}
\vspace{-7mm}
\caption{(\textit{Left} A.1-C.1) Continual classification accuracy over continual tasks. The number of novel classes introduced per task is in parenthesis. CUAL over-performs other methods in this challenging setting. The Oracle (gray) is fully supervised ER.(\textit{Right} A.2-C.2) Results varying the AL budget.}
\label{fig:plots}
\end{figure}

\vspace{-2mm}
Table \ref{fig:tables} (right) includes results for ablations of CUAL: 
\textit{AL-Top} selects samples with highest uncertainty scores (i.e. most-confidently novel samples) for AL rather than ambiguous samples; 
\textit{AL-Rand} refers to random sampling with the same supervision budget. These AL heuristic variations (Top and Rand) highlight the importance of querying Ambiguous samples instead of most novel samples as is customarily done.
Benefit of pseudo-labeling is clearly underscored by the sharp drop in performance in the \textit{only-AL} row in which pseudo-labeling is removed. In fact, AL only with an 8x higher labeling budget (\textit{only-AL(8x-budget)}) still does not achieve the same performance as that obtained by pseudo-labeling.
Note that in Fig 2 Table (left) we include additional results of the best-performing baseline ER, e.g. using Random or oneshot active labeling, akin to ablations of CUAL just described. 

\vspace{-7mm}
\begin{figure}[h]
    \scriptsize
    \centering
    \captionsetup[sub]{labelformat=empty,   
                   position = top}
    \subcaptionbox{}{
        \addtolength{\tabcolsep}{-0.43em}
        \begin{tabular}
        {c|cc|cc|cc|cc} 
        \toprule
        {} & \multicolumn{2}{c|}{Im21K} & \multicolumn{2}{c|}{Places} & \multicolumn{2}{c|}{Eurosat} & \multicolumn{2}{c}{Cifar100}\\ 
        {} & R50 & ViT & R50 & ViT & R50 & ViT & R50 & ViT\\
        \midrule 
        Oracle & 92.4 & 80.4 & 63.3 & 45.6 & 98.0 & 94.8 & 76.5 & 70.2 \\ 
        \midrule
        \textbf{CUAL(ours)} & \textbf{86.6} & \textbf{73.5} & \textbf{53.0} & \textbf{39.6} &  \textbf{95.8} & \textbf{94.1} & \textbf{65.0} & \textbf{65.7}\\ 
        ER-Ent & 76.1 & 62.8 & 41.7 & 31.3 &  91.0 & 87.3 & 56.2 & 57.4\\ 
        ER-Rand & 71.9 & - & 35.6 & - &   84.4 & - & 48.2 & - \\ 
        ER-Ent-oneshot & 75.3 & - & 39.4& - & 85.6 & - & 53.8 & -\\ 
        PseudoER-Ent  & 61.7 & 49.0 & 34.7 & 28.5 & 83.6 & 77.6 & 46.9 & 48.1\\
        CCIC  & 58.0 & - & 26.5 & - & 79.5 & - & 34.8 & -\\
        \midrule 
        \end{tabular}
    }
    \subcaptionbox{}{
        \raisebox{0.05\height}{
        \begin{tabular}[b]{p{2.1cm}p{0.5cm}p{0.5cm}p{0.5cm}p{0.5cm}}
        \toprule
        {Variations (R50)} & {Im21K} & {Places} & {Eurosat} & {Cifar100}\\ 
          \midrule
          CUAL(default) & \textbf{86.6} & \textbf{53.0} & \textbf{95.8} & \textbf{65.0}\\
          \midrule
          AL-Top & 75.0 & 48.4 & 88.8 & 62.9\\
          AL-Rand & 81.4 & 42.9 & 85.5 & 60.3\\
          AL-oneshot & 81.8 & 46.7 & 94.2 & 63.2 \\
          \midrule
          only-AL & 79.1 & 37.9 & 74.8 & 48.4\\
          only-AL-Rand$^*$ & 71.9 & 35.6 & 84.4 & 48.2\\
          only-AL(8x-budget) & 83.6  &  50.3 &  88.3 & 61.3\\
          \hline
        \end{tabular}
        }
    }
    \vspace{-2mm}
    \caption{Continual Classification Results, averaged over all tasks. \textbf{Left} - Default CUAL and baselines ("ent" equates to Entropy). \textbf{Right} - CUAL ablations. Refer to main text for discussion. $^*$see apx 4.3.}
    \label{fig:tables}
    \vspace{-2mm}
\end{figure}

\textbf{4. Conclusion:} In this work, we presented CUAL, a solution to the still under-explored field of continual active learning. To the best of our knowledge, CUAL is the first continual active learner to propose and demonstrate that querying based on the ambiguity of being novel or non-novel may be more informative than just selecting the samples that are most likely to be novel. Further, this work shows that assigning pseudo-labels with trustworthy confidence can significantly alleviate labeling costs without degrading continual AL performance. Overall, the proposed approach outperforms the baselines over multiple large-scale datasets and experimental variations. 
Yet, several challenges remain for continual active learning, which we hope to address in future work. One nontrivial example is how to continually query and update if distribution shifts of past classes (e.g. noise, illumination, etc) co-occur with novel classes.

{\small
\printbibliography
}

\section{Appendix}
\subsection{CUAL Methodology Details}

\subsubsection{Thresholds for Stopping the AL inner-loop and Pseudo-labeling} The inner-loop in CUAL is guided by two simple thresholds: (1) Threshold $T_{inner}$ "roughly" estimates if there are any possible novel-class samples in the unlabeled task input data pool and is controlled by a single hyper-parameter, the number of standard deviations above the mean of an in-distribution validation set (2 STDs in our experiments). If no samples are found to be above $T_{inner}$, we reach the stopping criterion for our iterations. Our in-distribution validation set is conventionally defined to include a portion (0.1\%) of the previous tasks' $k=1:t-1$ novelty predictions that were held-out at previous tasks, i.e. not used to update $A(x,t)$ parameters. Importantly, the same in-distribution validation set is used for all compared baselines in our results section, as is common practice in the OOD/novelty-detection literature \cite{rios2022incdfm, hsu2020generalized}; (2) Finally, Threshold $\alpha$ tunes pseudo-labeling selection and is set to $\alpha = 20\%$ highest $S^i(u)$ scores (most confident) from the test samples found above $T_{inner}$. These two thresholds are not highly sensitive.

\subsubsection{How to define Ambiguity} CUAL selects the most ambiguous samples at each AL inner-loop iteration, i.e. scores $S^i(u)$ which are neither too high or too low. Our mathematical formulation uses the threshold $T_{inner}$ defined in the previous section: we formulate ambiguousness as the inverse squared distance $\frac{1}{\|S^i(u)-T_{inner}\|^2}$ of scores to $T_{inner}$. Intuitively, this formula favors selecting samples that cannot be unambiguously predicted as either old or new since $T_{inner}$ represents this rough decision boundary. Active selection is stopped when the tiny labelling budget is exhausted. The only exception to this Ambiguity formulation is at the first iteration $i=0$ where we select homogeneously from samples above $T_{inner}$. This is the case because at $i=0$ only old classes are used to compute the score function, $S^0(u) = \min_{j\in C_{old}^t} FRE_j^0(u)$ and so ambiguity cannot be defined in the same way as for the remainder of iterations.

\subsubsection{Measuring per-class uncertainty in CUAL $S^i(u)$ formulation}
CUAL is agnostic to the elemental uncertainty metric used in its uncertainty scoring function ($S^i(u)$ Eq. 2 in section 2) as long as it can reliably estimate uncertainty w.r.t each novel class or old class. However, this is not an easy feat since many existing static uncertainty quantification approaches are not fully reliable \cite{ndiour2020probabilistic,rios2022incdfm}. As discussed in the main text, CUAL currently leverages the \textit{feature reconstruction error} (FRE) metric introduced in \cite{ndiour2020probabilistic} to build Eq 2. For each in-distribution class, FRE learns a PCA (principal component analysis) transform $\{\mathcal{T}_m\}$ that maps high-dimensional features $u$ from a pre-trained deep-neural-network backbone $g(x)$ onto lower-dimensional subspaces. During inference, a test-feature $u=g(x)$ is first transformed into a lower-dimensional subspace by applying $\mathcal{T}_m$ and then re-projected back into the original higher dimensional space via the inverse $\mathcal{T}_m^{\dagger}$. 
The FRE measure is calculated as the  $\ell_2$ norm of the difference between the original and reconstructed vectors:
\begin{equation}
\label{eq:FRE}
    FRE_m(u) = \|f(\mb{x})-(\mathcal{T}_m^{\dagger} \circ \mathcal{T}_m)u\|_2.
\end{equation}
Intuitively, $FRE_m$ measures the distance of a test-feature to the distribution of features from class $m$. If a sample does not belong to the same distribution as that $m$th class, it will usually result in a large reconstruction score $FRE_m$. FRE is particularly well suited for the continual setting since for each new class discovered at test-time, an additional principle component analysis (PCA) transform can be trained without disturbing the ones learnt for previous classes.

\subsubsection{Continual Model Update}

After the end of CUAL's inner loop, $i=I$ with $I$ corresponding to the last iteration, CUAL's short-term $A_s$ has accumulated a final set of actively labeled and pseudo-labeled novel samples. This final set is then used to update the continual agent $A(x,t)$: (1) obtain the final estimates of the PCA transformations corresponding to the novel classes, $\mathcal{T}_m^t, m \in C^{new}_t$, for task $t$ and to update long-term $A^{cl}$ (long term classification head of $A(x,t)$) via experience replay. From then on, in subsequent tasks, these novel classes $m, m \in C^{new}_t$ will no longer be flagged as ``novel'' and will be treated as ``old''. 

Here we expand upon the technique of "Experience Replay" (ER) \cite{rolnick2019experience, buzzega2021rethinking} used to update long-term $A^{cl}$ at the end of each task. Note that the same ER algorithm and hyper-parameters are used for baselines ER and PseudoER. In our implementation of ER a limited number of deep-embeddings belonging to old classes (from the frozen foundation model) must be stored in a Buffer of fixed size ($B_t={U_{old,k}^*, k:1:t-1}$). For a given task, samples $B_t$ are interleaved with those in $U_{new,t}^*$ to train $A(x,t)$ without catastrophic forgetting. After updating $A^{cl}$ the replay buffer itself must also be updated. We use a fixed-size memory buffer $B_t$ with the same building strategy as in \cite{rios2018closed}: a buffer of fixed size and prioritizing homogeneous distribution among classes. That is, an equivalent number of samples of each class are removed if room is required for new classes and the buffer is full. Formally, the ER training loss is defined in equation (\ref{eq:training_loss}), which includes cross-entropy loss, $L$, calculated over the set of actively-labeled new class samples ($new_{AL}$), pseudo-labeled (PL) samples ($new_{PL}$) and the buffer samples ($B_t$), respectively. 
We set the weights $\beta$, $\gamma$ and $\theta$ to 0.25, 0.25 and 0.5 respectively for a roughly even importance of old and new class samples.

\begin{equation}
\label{eq:training_loss}
    L_t = \beta L(new_{AL}) + \gamma L\left(new_{PL}\right) + \theta L(B_t)
\end{equation}

\subsection{Experimental Methodology}
\subsubsection{Implementation Details for CUAL and Baselines} 
We use two large-scale/foundation models as feature extraction backbones, kept frozen throughout CUAL and baselines' training: (1) Most results use ResNet50 \cite{he2016deep} unsupervisedly pre-trained on ImageNet1K via SwAV \cite{caron2020unsupervised}. We extract features from the pre-logit AvgPool layer of size 2048 as deep-embeddings. We also experimented with other feature extraction points \cite{ndiour2020probabilistic} but those under-performed w.r.t the pre-logit layer. (2) We also show results (Fig 2 Left Table) using ViTs16 \cite{alexey2020image} pretrained on Imagenet1K via DINO \cite{caron2021emerging}. For ViTs16 we tried several extraction points, e.g. head, last norm later, different transformer block outputs with different pool factors (e.g. 2,4). Best results were obtained with Block 9 features with pooling range of 2, yielding deep features of size 18816. Note that learning on frozen deep features is commonplace in vision CL and domain-adaptation fields \cite{rios2020lifelong,rios2022incdfm,evci2022head2toe}. It is theoretically based on the principle that low-level visual features from a large-scale/foundation frozen model are task nonspecific and do not need to be constantly re-learned. Rather, learning may happen upstream by utilizing the extracted deep features (at the last or inner-layers, or a combination thereof - an active research area) \cite{petrov2005dynamics,dhillon2019baseline,evci2022head2toe}. 

CUAL and baseline parameters are trained using the extracted deep features as outlined above. The classification heads are trained with ADAM \cite{kingma2014adam} and learning rate of 0.001. For long-term $A^{cl}$ we use a batchsize of 50 and an average of 20 epochs for each task during ER. CUAL $A_s^{cl}$ is trained at each inner-loop for an average of 5 epochs and mini-batch ranging from 10-20. As mentioned in the main text, we implement $A_s^{cl}$ as a fully-connected layer (we also tried 1-layer perceptrons ut with marginal performance gain). Baselines' long term classifier and CUAL's $A^{cl}$ are implemented as a one layer perceptron of size 4096 (also tested variations with marginal variations in results). The ER replay buffer is set to a size 5000 deep-embeddings for Places365 \cite{zhou2017places}, Imagenet21K-OOD \cite{ridnik2021imagenet21k} and 2500 for eurosat and cifar100. Lastly, baseline CCIC \cite{Boschini2022} was trained using same hyperparameters proposed by the authors and their open-source code.



\subsubsection{Datasets:} 
Since the employed large/foundation feature extractor were pretrained on Imagenet1K,
we evaluate CUAL on datasets that either do not contain class overlap with Imagenet1K (out-of-distribution w.r.t Imagenet1K \cite{2aa757143d6f46e2aba527d9e1a26aa5}), or curated them by excluding any overlapping classes. The exception is cifar100, which was included due to it being a very popular and widespread dataset. 
\begin{enumerate}
    \item \emph{Imagenet21K-OOD (Im21K-OOD) \cite{ridnik2021imagenet21k}}: We curated a subset of Imagenet21K containing the top-most populous 50 classes and that do not overlap with the classes present in Imagenet1K. We use a random set of 500 samples from each of the 50 classes. Because Imagenet21K is a superset of Imagenet1K, by excluding any overlapping class we guarantee orthogonality in our curated subset. We will release the full list of images chosen in this curation for reproducibility. 

    \item \emph{Places365-OOD (Places) \cite{zhou2017places,huang2021mos}}: is a subset of Places365 also originally curated by \cite{huang2021mos} to contain 51 "environment" categories orthogonal to Imagenet1K, containing a total of 9822 images. It has also been used as a test OOD dataset in \cite{huang2021mos,ming2022delving,sun2021react} with respect to Imagenet1K. 

    \item \emph{Eurosat \cite{helber2019eurosat}}: An RGB dataset of 10 classes and 27K images of Sentinel-2 satellite images, which is also orthogonal to Imagenet1K.

    \item \emph{Cifar100-Superclasses (Cifar100) \cite{cifar100}}: We use the super-label granularity of Cifar100 dataset. This totals 20 labels (super) and 50K images. While Cifar100  is not orthogonal to Imagenet1K, we decided to showcase its results since it is a widespread dataset in CL. 
\end{enumerate}

\subsubsection{Baselines} 

As discussed, we use a range of methods for comparison: (1) CCIC \cite{Boschini2022} leverages experience replay, a MixMatch \cite{berthelot2019mixmatch} derived label sharpening technique and nearest neighbor classification to solve semi-supervised CL but does not differentiate between old and new classes which may be unlabeled. we used the author's publicly released code to generate results. CCIC, similar to other semi-supervised CL models, assumes that new and old classes will not co-occur (discussed in Section 1) which ultimately explains its performance degradation when the oracle assumption (of only novel classes present in an unlabeled pool) is no longer valid, as in our experimental scenario; (2) ER \cite{rolnick2019experience, buzzega2021rethinking}, originally proposed for supervised CL is adapted in a similar fashion to continual active learning work \cite{vu2023active} to only use actively labeled samples (as embeddings) for replay; (3) We also adapt PseudoER \cite{pseudoDER} to continual AL by further incorporating pseudo-labeling for high confidence unlabeled samples in addition to actively labeled samples to be used for training. In both ER and PseudoER, we utilize the cumulative classification entropy as an uncertainty score to actively-label and Pseudo-Label (PseudoER). Similar to CUAL, we actively label ''ambiguous'' samples according to the same formula as outlined in appendix 4.1.2 for superior results, then sampling according to the TOP heuristic (see section 3 discussion). We also tested with other common uncertainty metrics such as margin \cite{ren2021survey} but with inferior results. 

\subsection{Additional Results}

\begin{figure}[h]
\centering
\includegraphics[width=7cm]{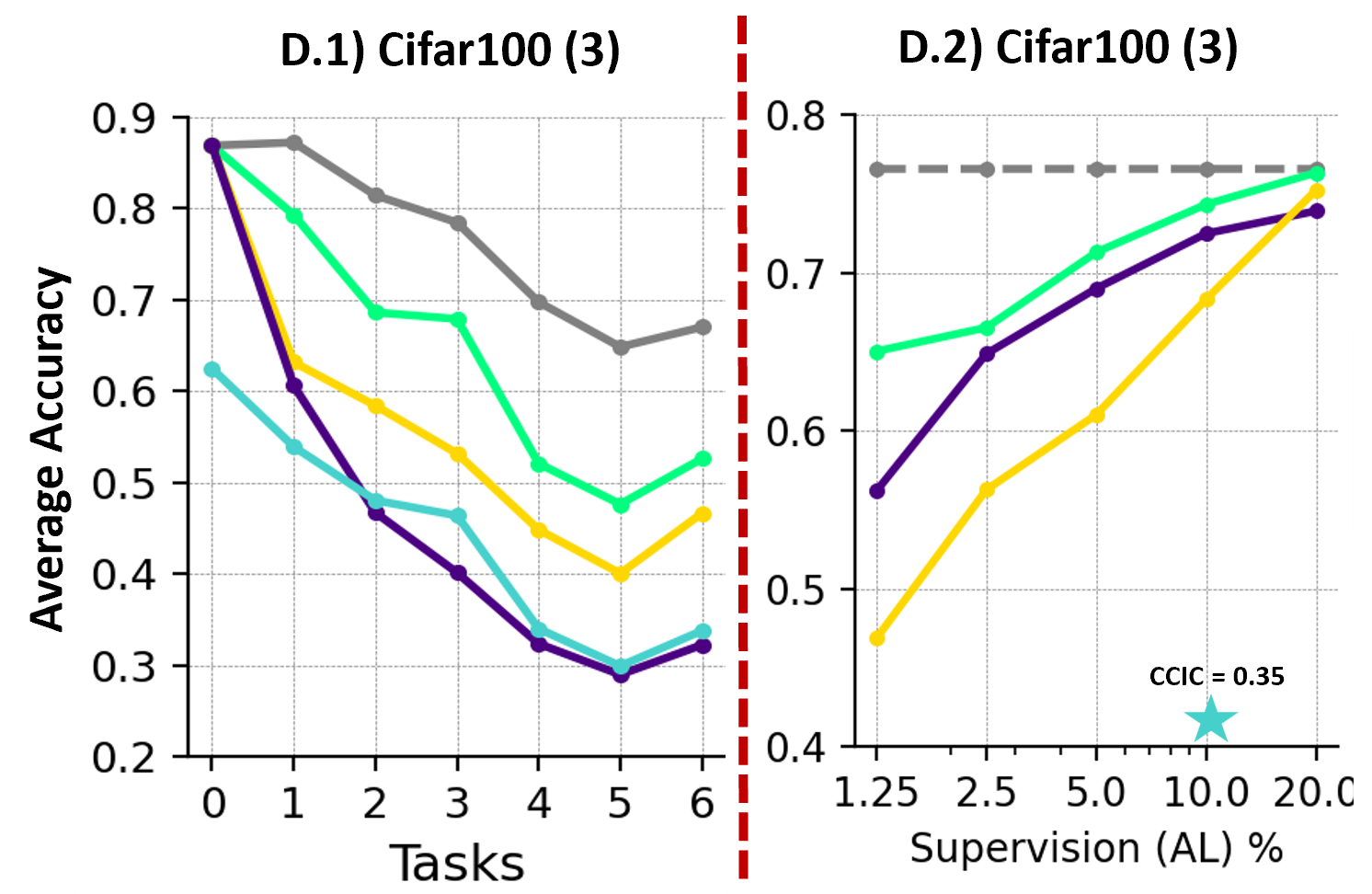}
\caption{Results for Cifar100; (Left D.1) Continual classification accuracy over continual tasks. The number of novel classes introduced per task is in parenthesis. Al budget is 1.25\% as outlined in section 3.1. CUAL over-performs other methods in this challenging setting. The Oracle (gray) equates to classic fully supervised experience-replay.(Right D.2) Varying the Active supervision budget and it's effect on CUAL and AL baselines.}
\label{fig:cifar}
\end{figure}

\begin{table}[h]
    \footnotesize
    \centering
    \begin{tabular}{ccccc}
         \toprule
          Class-inc & 1 & 3 & 5 & 7\\
          \midrule
         Oracle &  92.4 & 92.2& 92.4&91.4\\
         \midrule
         \textbf{CUAL(ours)} & \textbf{90.7} & \textbf{88.3}& \textbf{86.6}&\textbf{83.6}\\
         ER-ent & 77.9 & 74.2& 76.1& 78.2\\
         PseudoER-ent & 65.5 & 55.9& 61.7& 59.6\\
         CCIC & 63.1 & - & 58.0 & - \\
         \hline
    \end{tabular}
    \caption{Averaged continual classification performance when varying class increments per task. Results for Im21K-OOD with AL budget of 2.5\%.}
    \label{tab:OneClass}
\end{table}

Result clarification for Fig 2 right table (ablations): The ablation only-AL-rand is equivalent to ER-rand since sampling is done randomly during the inner loop. In other words, it does not use either entropy or Eq \ref{eq:uncertainty-uncanny}
to guide AL. Furthermore, all of the only-AL ablations do not use pseudo-labeling.

\end{document}